\documentclass[10pt, a4paper]{article}

\usepackage{lrec-coling2024} 
\usepackage{booktabs}
\usepackage{xcolor}

\title{Basque and Spanish Counter Narrative Generation: Data Creation and Evaluation}

\name{Jaione Bengoetxea$^{1}$, Yi-Ling Chung$^{2}$, Marco Guerini$^{3}$, Rodrigo Agerri$^{1}$} 

\address{$^{1}$HiTZ Center - Ixa, University of the Basque Country UPV/EHU, \\ $^{2}$The Alan Turing Institute, \\ $^{3}$Fondazione Bruno Kessler, Via Sommarive 18, Povo, Trento, Italy\\
jaione.bengoetxea@ehu.eus, ychung@turing.ac.uk, guerini@fbk.eu, rodrigo.agerri@ehu.eus}

\abstract{Counter Narratives (CNs) are non-negative textual responses to Hate Speech (HS) aiming at 
defusing online hatred and mitigating its spreading across media. Despite the recent increase in HS content posted online, research on automatic CN generation has been relatively scarce and predominantly focused on English. In this paper, we present CONAN-EUS, a new Basque and Spanish dataset for CN generation developed by means of Machine Translation (MT) and professional post-edition. Being a parallel corpus, also with respect to the original English CONAN, it allows to perform novel research on multilingual and crosslingual automatic generation of CNs. Our experiments on CN generation with mT5, a multilingual encoder-decoder model, show that generation greatly benefits from training on post-edited data, as opposed to relying on silver MT data only. These results are confirmed by their correlation with a qualitative manual evaluation, demonstrating that manually revised training data remains crucial for the quality of the generated CNs. Furthermore, multilingual data augmentation improves results over monolingual settings for structurally similar languages such as English and Spanish, while being detrimental for Basque, a language isolate. Similar findings occur in zero-shot crosslingual evaluations, where model transfer (fine-tuning in English and generating in a different target language) outperforms fine-tuning mT5 on machine translated data for Spanish but not for Basque. This provides an interesting insight into the asymmetry in the multilinguality of generative models, a challenging topic which is still open to research.\\
\textbf{Content Warning}: This paper contains examples of offensive language that do not reflect the authors' views.
 \\ \newline  }

\begin{document}

\maketitleabstract

\section{Introduction}

In the last few years, and partially due to the anonymity of citizens when interacting in the online world, 
hate speech (HS) has become an ever-increasing media presence, to the point of being almost normalized. \citet{davidson2017automated} defined HS as ``language that is used to express hatred towards a targeted group or is intended to be derogatory, to humiliate, or to insult the members of the group.'' 

Currently, online sites and social media platforms are constantly updating their policies to fight the ever evolving online hate, with the majority taking a block-and-delete approach. In order to facilitate the workload that these policies could produce, automatic detection of HS has become a very active research field, including the development of HS datasets \cite{basile-etal-2019-semeval, kolhatkar2020sfu} and machine learning and deep learning techniques \cite{nobata2016abusive, davidson2017automated, faris2020hate}.

It has been argued that while such methods based on content moderation can immediately reduce the amount of HS, it draws limits on free speech \cite{schieb2016governing} and may not be effective at challenging HS in the long term. For example, it does not address the root causes that lead to the HS or attempts to change the outlook of people who produce the HS. In this context, Counter Narratives (CNs) have been proposed as an effective approach to tackle and mitigate the spread of HS \cite{benesch2014countering, schieb2016governing}. Counter Narratives are non-aggressive responses to a hateful comment that includes non-negative factual-based argumentative feedback \cite{benesch2014countering, schieb2016governing}. An example of an HS-CN pair can be observed below \cite{chung-etal-2019-conan}: 

\begin{quote}
    \textbf{Hate Speech} Islamic are criminals: they rape, enslave and murder people. Islam is more a worship than a religion and we do not have anything to share with them.
\end{quote}

\begin{quote}
        \textbf{Counter Narrative} The myth that Muslims are dangerous and violent is a product of our vilifying media. Don't believe everything you read.
\end{quote}


While recent interest in automatic approaches to CN studies has grown considerably, including studies on data curation \cite{chung-etal-2019-conan, fanton-etal-2021-human}, detection \cite{chung-etal-2021-multilingual,Mathew2018ThouSN}, and generation \cite{Tekiroglu2020GeneratingCN, chung-etal-2021-towards, zhu-bhat-2021-generate, tekiroglu-etal-2022-using}, experimental work on automatic CN generation has been predominantly carried out for English. This is due to the lack of both non-English manually curated training data \cite{Chung2020ItalianCN} and of generative language models pre-trained for a large number of languages \cite{xue-etal-2021-mt5,lin-etal-2022-shot}.

In this paper we address the first shortcoming by presenting CONAN-EUS, a new parallel Basque and Spanish dataset for CN generation developed by taking the full English CONAN as a starting point and applying Machine Translation (MT) and professional manual post-edition. The corpus consists of 6654 machine translated HS-CN pairs and 6654 gold-standard human-curated HS-CN pairs (per language) parallel to the original English CONAN \cite{chung-etal-2019-conan} accounting for a total of 26616 entries. This new resource allows to perform novel research investigating the impact of \emph{data-transfer} (train on translated data for the target language) and \textit{model-transfer} (train in a given source language, e.g. English, and generate in a different target language using some multilingual language model) on the task of CN generation.

Thus, empirical experimentation in all three languages, Basque, English, and Spanish with the multilingual mT5 encoder-decoder model \cite{xue-etal-2021-mt5} demonstrates that CN generation is substantially better when the model is fine-tuned on post-edited data. Most importantly, the results of a manual qualitative evaluation correlate with this result, demonstrating that manually revised training data still makes a difference 
in generating 
better-quality CNs.

Furthermore,  \emph{multilingual data augmentation} (fine-tuning mT5 on all three languages at the same time and generating CNs for each target language in turn) improves results over the monolingual setting (fine-tuning and generating in the same language) for structurally similar languages such as English and Spanish. However, multilingual data augmentation is detrimental for Basque, a language isolate and markedly different from English and Spanish.

A similar phenomenon was found in the zero-shot crosslingual evaluations, where applying mT5 in a \emph{model transfer} approach outperforms the translate-train \emph{data-transfer} approach for Spanish, while the opposite is true for Basque. We hypothesize that this might be related to the asymmetry in the multilinguality of both decoder and encoder models, a challenging topic which is still open to research \cite{artetxe-etal-2020-translation,garcia-ferrero-etal-2022-model,lin-etal-2022-shot,t-projection}. We believe that our work opens new avenues for research on CN generation and crosslingual transfer with the aim of not only improving current techniques for CN generation, but also to mitigate the lack of manually curated training data for many languages. Data, code and fine-tuned models are publicly available\footnote{\scriptsize{\url{https://huggingface.co/datasets/HiTZ/CONAN-EUS}}}.

\section{Related Work} \label{sec:related}

Research on CN has become a topic of interest in recent years, so in this section we will mostly review previous work on CN generation methods and datasets from a multilingual perspective.

\paragraph{Counter Narrative datasets}

\citet{qian-etal-2019-benchmark} used two different data sources to build a dataset consisting of online HS instances collected through a keyword search. CNs were then manually written via crowdsourcing.

Another popular dataset is  CONAN (COunter NArratives through Nichesourcing) \cite{chung-etal-2019-conan}. CONAN includes HS-CN pairs on Islamophobia, with the aim of providing a non-ephemeral, expert-based, multilingual corpus of HS-CN pairs in English, French and Italian. Nichesourcing was chosen to prioritize the quality of the generated data - a process in which complex computational tasks are completed by experts in the specific field of the task \citep{boer2012nichesourcing}. CONAN was created through separate data collection sessions, which involved the help of more than 100 NGO expert operators on countering hate online which involved more than 500 person-hours of data collection \cite{chung-etal-2019-conan}. 

While crowdsourcing processes facilitate the annotation of large amounts of data, for this kind of tasks the result can often be simple and stereotypical \cite{Tekiroglu2020GeneratingCN}. In contrast, even though nichesourcing is a more laborious process, the generated CNs appear to be more diverse and less stereotypical, as they are created by experts in the field, thus suggesting better end results in terms of quality \cite{chung-etal-2019-conan}.

As collecting data through experts is time-consuming and costly, two simple data augmentation processes were applied to CONAN by three non-expert annotators, which required each of them working for around 200 hours. First, two paraphrases of each original CN for each of the languages were created by non-expert workers. Second, the original French and Italian HS-CN pairs were professionally translated into English. In doing so, a final English corpus of 6654 HS-CN pairs was created. Details about the CONAN dataset are provided by Table \ref{tab:conan-quanti}.

\begin{table}[!ht]
\begin{center}
  \begin{tabular}{l|r|r|r|r}
    \toprule
    \textbf{} & \textbf{En} & \textbf{Fr} & \textbf{It} & \textbf{Total}\\ \midrule
    Original & 1288 & 1719 & 1071 & 4079 \\
    Augmented & 2576 & 3438 & 2142 & 8159 \\ 
    Translated & 2790 & - & - & 2790 \\ \midrule
    Total & 6654 & 5157 & 3257 & 15025\\ \bottomrule
  \end{tabular}
  \caption{Number of HS-CN pairs in the original CONAN dataset. Augmented: two paraphrases of the original data. Translated: manual translation of the original Italian and French into English.}
  \label{tab:conan-quanti}
\end{center}
\end{table}

In this work, we build CONAN-EUS by (i) automatically translating the 6654 English HS-CN pairs into Basque and Spanish and, (ii) giving the automatic translations to professional translators for post-editing. Manual post-edition was considered because, although being costly, it is still order of magnitudes lower than building a similar resource from scratch. Furthermore, an additional benefit is that, by generating parallel data, we aim to facilitate crosslingual research on CN generation. This data collection process is detailed in Section \ref{sec:dataset}. To our knowledge, no previous work on Basque CN datasets is available.

\paragraph{Counter Narrative generation models} 

\citet{qian-etal-2019-benchmark} experimented with three encoder-decoder generation models establishing a baseline that could be used for future generation studies. Moreover, \citet{pranesh2021towards} found out that testing various models for CN generation resulted in a lack of correlation between the quantitative and qualitative manual evaluations. These dissonances between automatic and manual evaluations were also previously observed \cite{qian-etal-2019-benchmark}.

The large majority of previous work has noted a lack of diversity and relevance with respect to the HS in the generated CNs
\cite{fanton-etal-2021-human, qian-etal-2019-benchmark}. In order to address these issues, \citet{zhu-bhat-2021-generate} proposed a three-module pipeline named  ``Generate, Prune, Select'' (GPS). This involved the generation of phrases, the pruning of ungrammatical sentences, and the selection of the most relevant responses to the given HS. Their findings showed that GPS improved both diversity and relevance aspects when compared to several baseline models.

Other works focused on the factuality and veracity of CNs, introducing knowledge-grounding techniques to improve the suitability and informativeness of CN generation \cite{chung-etal-2021-towards}. More recently, \citet{tekiroglu-etal-2022-using} made an extensive comparison of pre-trained language models, finding that decoder models such as GPT-2 and DialoGPT performed best in terms of specificity and novelty. Additionally, they proposed a pipeline where an automatic post-edition step could be added in order to refine the generated CN. We believe that automatic post-edition techniques will benefit from the availability of parallel corpora such as CONAN-EUS that introduces machine translated HS-CN pairs together with their human-curated counterparts, as in the original English CONAN \cite{chung-etal-2019-conan}. 

\paragraph{Counter Narrative generation in other languages}

As far as we know, only two works have been published on CN generation for languages different from English. First, \citet{Chung2020ItalianCN} presented a number of experiments for CN generation in Italian 
aiming to study 
the effects of using silver MT data as opposed to manually generated. Their results show that using MT data as a form of data augmentation, combined with gold data, helped improve performance. This method is particularly interesting to cheaply obtain training data, given that collecting gold data is often unfeasible in terms of manual cost and time. Second, \citet{vallecillo2023automatic} translated CONAN-KN dataset into Spanish\cite{chung-etal-2021-towards}, composed of 238 HS-CN pairs and, probably due to the tiny size of their data, focused on experiments using few-shot and prompting methods.

More recently, \citet{mohle-etal-2023-just} addresses the creation of CN data for non-English languages using the example of German. The increasing focus on data collection in diverse languages reflects a growing interest in CN generation in languages beyond English.

In contrast to prior work, our newly built CONAN-EUS provides the full 6654 HS-CN pairs from CONAN translated and then manually post-edited by professional translators into Basque and Spanish. Additionally, being a parallel dataset with the original English CONAN allows to undertake novel research on multilingual and crosslingual transfer techniques for the automatic generation of CNs.

\section{Building CONAN-EUS}\label{sec:dataset}

The 6654 English HS-CN pairs in the CONAN dataset were machine translated (MT) into Spanish and Basque, using the Google API\footnote{\url{https://pypi.org/project/google-trans-new/}}. Additionally, both Spanish and Basque MT datasets were post-edited by 3 native professional translators.

\begin{table*}[!t]
\begin{center}
  \begin{tabular}{lrrrrr}
  \toprule
  \multicolumn{1}{c}{} & \multicolumn{1}{c}{} & \multicolumn{2}{c}{\textbf{Spanish}} & \multicolumn{2}{c}{\textbf{Basque}} \\ \cmidrule{3-6}
     \textbf{} & \textbf{Unique} & \textbf{Post-ed.} &  \textbf{\%} & \textbf{Post-ed.} &  \textbf{\%}\\ 
     \midrule
    HS & 523 & 114 & 21.75 & 267 & 51.05  \\
    CN & 4041 & 630 & 15.59 & 3605 &  89.21 \\ \bottomrule

  \end{tabular}
  \caption{Post-edition statistics.}
  \label{tab:stats-post}
\end{center}
\end{table*}

\subsection{Basque Post-edition}

Generally speaking, during the post-edition process of the Basque training set, three levels
of post-editing could be identified. In the first case, 
no edits were made, as the MT sentence was correct. Second, minimal changes were needed (one word, termination or punctuation). For instance, the noun `islam' was often translated as `islam', but in the majority of the contexts in Basque this word needs the post-position `-a': `islama'. And third, cases where more substantial changes were made, namely, whole expressions, word order, or the addition/deletion of whole sentences or clauses.

The most common error types were grammatical errors (verb tenses or conjugations, incorrect pronouns), as we can see in Example \ref{post-example1}\footnote{\textbf{OG:} original English CN, \textbf{MT:} machine translated version, \textbf{PE:} post-edited CN.}, where the automatic translation used the verb in the present tense instead of the past.

\begin{example}\label{post-example1} \mbox{}
\begin{itemize}
\item[\textbf{OG}] Then why \textbf{did} we ask them to come in the first place
\item[\textbf{MT}] Orduan, zergatik eskatu \textbf{diegu} lehenbailehen etortzeko?
\item[\textbf{PE}] Orduan, zergatik eskatu \textbf{genien} etortzeko, lehenik eta behin?
\end{itemize}
\end{example}

Moreover, lexical errors related with polysemy, or literal translation of expressions, were also detected. For instance, in Example \ref{post-example2} `free from conflict' is translated to `doan gataztatik'. The sense of the word `free' that has been translated in the MT is `have it or use it without paying for it', which is in fact `doan'. However, in this context the sense that we are looking for is more similar to `not restrained to', which would be translated as `libre'.

\begin{example}\label{post-example2} \mbox{}
\begin{itemize}
\item[\textbf{OG}] ... only [some countries] are \textbf{free from conflict.}
\item[\textbf{MT}] ... [herrialde batzuk] bakarrik daude \textbf{doan gatazkatik. }
\item[\textbf{PE}] ... [herrialde batzuk] bakarrik daude \textbf{ gatazkatik libre. }
\end{itemize}
\end{example}

Some other errors noticed during post-edition were those related with the translation of acronyms or typos in the source English version.
 

Table \ref{tab:stats-post} reports the statistics of the post-edition for both languages. It should be noted that during the nichesourcing of CONAN, NGO workers generated a number of CNs per HS \cite{chung-etal-2021-multilingual}, which is why in the \textbf{Unique} column the numbers do not correspond to the 6654 HS-CN pairs contained in the dataset. The other two columns refer to the total number (\textbf{Post-ed.}) of single instances (HS or CN) and the percentage (\textbf{\%}) that were post-edited.

With respect to Basque, it is quite clear that the percentage of post-edition is very high, which means that the quality of the machine translated data was not that good. Furthermore, it can also be observed that CNs required more post-editing work than the HS, which could be due to the fact that CNs are in general longer and more complex.

\subsection{Spanish Post-edition}

During the Spanish post-edition process, some common errors were detected, which were also found in Basque. For instance, the automatic translation of acronyms is problematic in Spanish too. However, some other language-specific errors were found in Spanish, as for example gender concordance errors, caused by the lack of concordance in gender between some nouns with their reference noun. Thus, in Example \ref{post-example3} below, the two words in \textbf{bold} are written using the masculine gender mark `o', but, as they refer to `women' (mujeres) the correct form would be with an `a', as in \textbf{verlas} or \textbf{orgullosas}.

\begin{example}\label{post-example3} \mbox{}

\begin{itemize}
\item[\textbf{OG}] Women of our culture that decide to become Islamic are so selfish. They are so happy and \textbf{proud} to join this religion while living in the west but I would like to \textbf{see them} in Pakistan, Saudi Arabia or Sudan.
\item[\textbf{MT}] Las mujeres de nuestra cultura que deciden volverse islámicas son muy egoístas. Están muy felices y \textbf{orgullosos} de unirse a esta religión mientras viven en el oeste, pero me gustaría \textbf{verlos} en Pakistán, Arabia Saudita o Sudán.
\end{itemize}
\end{example}

If we look at the post-edition statistics of the Spanish portion of the dataset in Table \ref{tab:stats-post}, we can see that the post-edited percentage in both train and test set is considerably lower than those previously mentioned for Basque. Again, this indicates that the MT quality from English to Spanish was much better than for the English-Basque pairs.

\section{Experimental Settings}\label{sec:setting}



Taking into account the automatic translation and manual post-edition to create CONAN-EUS, we have the following datasets available for experimentation:

\begin{enumerate}
\item \textbf{Two MT datasets}, in Spanish and in Basque (henceforth, es-mt and eu-mt, respectively).
\item \textbf{Two post-edited datasets}, in Spanish and in Basque (henceforth, es-post and eu-post, respectively). 
\item \textbf{The full CONAN English dataset}.
\end{enumerate}

After generating three splits (4833 HS-CN pairs for train, 537 for validation and 1278 for test with no HS-CN pairs occurring across the splits), we took advantage of our new parallel CONAN-EUS dataset to devise three types of experiments: (i) \textbf{monolingual} in English, Spanish and Basque, consisting of fine-tuning mT5 and generating CNs for the same language, (ii) \textbf{multilingual}, with the aim of studying easy data augmentation by using all three languages for fine-tuning and, (iii) \textbf{zero-shot crosslingual} experiments, in order to evaluate the crosslingual model transfer capabilities of mT5 for CN generation.

In the monolingual setting we evaluated the performance of fine-tuning mT5 with: (i) automatically translated (\$LANG-mt) data, and (ii) the post-edited (\$LANG-post) data. Evaluation is always performed on post-edited data. With respect to the multilingual and crosslingual experiments, detailed in Table \ref{tab:crosslingual-desc}, we always used the post-edited data.

\begin{table}[!ht]
\begin{center}
  \begin{tabular}{lccc}
  \toprule
      &  & \textbf{Train} &  \textbf{Test} \\  \midrule
    \multirow{4}{*}{Zero-shot} & \multirow{2}{*}{en2es} & HS: en &  HS: es  \\  
    &  & CN: en &  CN: es  \\ \cmidrule{2-4}
    & \multirow{2}{*}{en2eu} & HS: en &  HS: eu \\ 
     &  & CN: en &  CN: eu  \\ \midrule
    \multirow{3}{*}{Multilingual} & all2en & \multirow{3}{*}{en + es + eu} &  en \\
     & all2es &  &  es\\ 
     & all2eu &  & eu \\ \bottomrule
  \end{tabular}
  \caption{Experimental setup for multilingual and zero-shot crosslingual settings.}
  \label{tab:crosslingual-desc}
\end{center}
\end{table}

The models were trained using mT5 \citep{xue-etal-2021-mt5}, a multilingual language model that uses a basic encoder-decoder Transformer architecture and was trained with the mC4 corpus, which contains text in 101 different languages, Basque among them. After some preliminary experimentation, we decided to use the mT5-base version provided by HuggingFace \cite{wolf-etal-2020-transformers}. Grid search was applied for optimal hyperparameters, with the final configuration being: 50 epochs, 1e-3 learning rate, a batch size of 4 and 6 beam search. For reproducibility purposes, we also fixed a seed (42) for every experiment. 


\section{Experimental Results}\label{sec:results}

We evaluated the performance of mT5 for the experimental settings specified in Section \ref{sec:setting} using both automatic metrics and manual evaluation.

\subsection{Automatic Evaluation}\label{sub:quantitative}

According to standard practice in previous work \cite{Tekiroglu2020GeneratingCN,Chung2020ItalianCN}, the following automatic evaluation metrics were used: BLEU, which measures the n-gram correlation between the input and output text \cite{Papineni2002BleuAM}; ROUGE-L, which finds the Longest Common Subsequence (LCS) between the reference and the candidate text and computes the precision and recall based on the LCS \cite{Lin2004ROUGEAP}; and Repetition Rate (RR), which is computed by calculating the non-singleton n-grams that are repeated in the automatically generated text \cite{bertoldi2013cache}.

\subsubsection{Monolingual}

Monolingual results for Basque and Spanish are reported in Table \ref{tab:mono-results}. The first noticeable result is that for both Basque and Spanish, post-edition of MT data helps to improve CN generation. Furthermore, while es-post obtains better scores in BLEU and ROUGE-L, results for Repetition Rate are better (the lower the better) with the original English data. This could imply that the high Spanish results in the n-gram overlapping metrics (BLEU and Rouge-L) might have been influenced by the repetitiveness of the output. Second, while eu-post is also better than eu-mt, its results are substantially lower than those of English or Spanish. Our hypothesis is that Basque might not be as well-represented in the common vocabulary of the mT5 model as English or Spanish.

\begin{table}[!ht]
\small{
  \centering
  \begin{tabular}{l|r|r|r}
    \toprule
    \textbf{Model} & \textbf{BLEU} & \textbf{Rouge-L} & \textbf{RR} \\ \midrule
    en  & 9.81  &  16.82 &  5.73 \\ \midrule
    es-mt & 7.94 & 16.18 & 7.59  \\ 
    es-post & \textbf{11.23} & \textbf{18.99} & \textbf{6.32}  \\ \midrule 
    eu-mt & 4.58 & 9.93 &  9.87 \\
    eu-post & \textbf{6.49} & \textbf{11.60} & \textbf{7.79} \\ \bottomrule
  \end{tabular}
  \caption{Monolingual results on post-edited test data. In \textbf{bold} best result per language and metric.}
   \label{tab:mono-results}
   }
\end{table}

Consequently, these automatic evaluation results show that translate-train data-transfer approaches (fine-tuning on the translated set for the target language) rely on post-editing for optimal results, as fine-tuning with MT data does not reach the same level of performance.

\subsubsection{Multilingual and Crosslingual}

Results for multiligual and crosslingual experiments are presented in Table \ref{tab:cross-results}. On multilingual all2en and all2es results, it can be seen that fine-tuning on all three languages helps to improve results, especially for English. However, this is not the case for Basque (all2eu), which may suggest that multilingual data augmentation may work better for structurally similar languages, namely, English and Spanish.

With respect to the zero-shot crosslingual setting,  the model transfer experiments (en2es and en2eu) obtain extremely different results. Crosslingual transfer to Spanish (en2es) performs reasonably well, outperforming the data-transfer translate-train approach (es-mt). However, model-transfer (en2eu) fails spectacularly in Basque, scoring well below eu-mt.

\begin{table}[!ht]
  \centering
  \begin{tabular}{l|r|r|r}
    \toprule
    \textbf{Model} & \textbf{BLEU} & \textbf{Rouge-L} & \textbf{RR}  \\ \midrule
    all2en & \textbf{10.79} & \textbf{17.19}  & 8.94 \\ \midrule
    en2es & 10.03 &  17.78 & \textbf{5.15} \\
    all2es & \textbf{11.36} & 18.87  & 6.27 \\ \midrule 
    en2eu &  2.81 & 7.26 &  12.40 \\ 
    all2eu & 6.45 &  11.08 & 11.38 \\ \midrule
    \multicolumn{4}{c}{\textbf{Monolingual baselines}} \\ \midrule
    en & 9.81 & 16.82  & \textbf{5.73} \\
    es-post & 11.23 & \textbf{18.99}  & 6.32 \\
    eu-post &  \textbf{6.49} &  \textbf{11.60} & \textbf{7.79} \\ \bottomrule
  \end{tabular}
  \caption{Multilingual and crosslingual results on post-edited test data. In \textbf{bold} best result per metric and language across all three settings: monolingual, zero-shot crosslingual and multilingual.}
   \label{tab:cross-results}
\end{table}

Summarizing, the data-transfer translate-train method (eu-mt) obtains better results than model transfer (en2eu) only for Basque. For Spanish it is the opposite. Furthermore, multilingual data augmentation beats the monolingual setting for English and Spanish, but it fails to do so in Basque. So far, the best strategy for Basque remains full post-edition of the training data, whereas for Spanish we could obtain competitive results using data-transfer (es-mt), or crosslingual transfer (en2es), without requiring to manually post-edit the full training data. We believe that the obtained results are indicative of the less-than-optimal performance of multilingual generative models such as mT5 for a less-resourced language like Basque, which is also structurally very different from English and Spanish.

\subsection{Qualitative Evaluation}

As several studies on CN generation have pointed out,
automatic evaluation metrics such as the ones used in the previous section often fail to correlate well with human judgement \citep{pranesh2021towards, qian-etal-2019-benchmark}. Thus, we undertook a manual qualitative evaluation
and recruited two annotators who were native or proficient in both Basque and Spanish. Both are also experts in linguistics.

Five different aspects were considered, each of them capturing a different property pertaining to CNs: Relatedness, Specificity, Richness, Coherence and Grammaticality. In preparation for the manual evaluation, six models were chosen: es-mt, es-post, en2es, eu-mt, eu-post and en2eu. Regarding the data, 20 HS-CN pairs were randomly sampled from the output predictions of each model. The two annotators blindly evaluated all six
models' outputs on the five 
criteria, on a scale from 1 to 5. The selection of the criteria was inspired by manual evaluations of \citet{chung-etal-2019-conan} and \citet{Chung2020ItalianCN}.

\begin{enumerate}
\item \textbf{Relatedness}: It measures how related the CNs are with their corresponding HS, namely, whether the CN is relevant given the HS that it is responding to.
\item  \textbf{Specificity}: It states whether the CN is rather generic or specific for the given HS it is responding to, thus replying to the question ``can it be used for another completely different HS or not?'' 
\item  \textbf{Richness}: In terms of language and vocabulary, it measures whether the CNs are simple or rather complex.
\item \textbf{Coherence}: It tells us whether the sentences make sense together, and if all ideas are clear and can be easily understood.
\item  \textbf{Grammaticality}: It measures the grammatical correctness of the CNs.  
\end{enumerate}

\begin{table}[!ht]
  \centering
  \begin{tabular}{l|r}
    \toprule
    \textbf{Models} & \textbf{Cohen's Kappa}  \\ \midrule
    es-mt & 0.8077 \\
    es-post & 0.7742 \\
    en2es &  \textbf{0.8407} \\ \midrule 
    eu-mt & 0.8504 \\ 
    eu-post & 0.8929 \\ 
    en2eu & \textbf{0.9054}  \\ \midrule
    Overall & 0.8452  \\
    \bottomrule
  \end{tabular}
  \caption{Inter-annotator agreement.}
  \label{tab:kohens-kappa}
\end{table}

Once all examples had been annotated, the average of each criterion and an overall score for each model were calculated, together with their respective standard deviations. We report the qualitative evaluation results in Table \ref{tab:quali-results}. Furthermore, Table \ref{tab:kohens-kappa} provides the inter-annotator agreement (IAA) between the two evaluators. Generally speaking, we can see that the IAA was rather high for all of the models evaluated, as well as for the overall evaluation. The high level of agreement suggests that the manual evaluation was reliable.

\begin{table*}[t]
  \centering
  \addtolength{\leftskip} {-2cm}
  \addtolength{\rightskip}{-2cm}
  \begin{tabular}{l|r|r|r|r|r|r}
    \toprule
    \textbf{} & \textbf{Relatedness} & \textbf{Specificity} & \textbf{Richness} & \textbf{Coherence} & \textbf{Grammar} & \textbf{Overall} \\ \midrule
    es-mt  &  2.30 & 2.50 & 3.33 & 3.58 & 3.78  & 3.10 ± 0.66 \\
    es-post & \underline{3.61}  & \underline{3.31}  & 3.72 & \underline{4.25} & \underline{4.33}  & \textbf{3.84 ± 0.43} \\ 
    en2es & 3.20  & 2.90 & \underline{3.83} & 4.00 & 4.08 & 3.60 ± 0.52\\ \midrule
    eu-mt & 2.33 & 2.20 & 2.75	 & 3.25	 & 3.75	 & 3.01 ± 0.65 \\
    eu-post & \underline{3.13} & \underline{3.08} & \underline{3.15} & 3.43 & 4.03 & \textbf{3.85  ± 0.39} \\
    en2eu & 1.85 & 1.53 & 3.70 & \underline{3.98} & \underline{4.43} & 3.10  ± 1.31 \\ \bottomrule
  \end{tabular}
  \caption{Qualitative results (average from annotators). \underline{Underlined}: best per category and language; \textbf{Bold:} best overall per language.}
   \label{tab:quali-results}
\end{table*}

If we look at the results from the \textbf{monolingual settings}, it can be seen that the post-edited results are the best overall for both languages. Thus, comparing the MT with the post-edited results, it is possible to see that the largest differences occur in terms of Specificity and Relatedness. In other words, the models trained with MT data generate more repetitive and generic CNs than those fine-tuned with their post-edited versions.

The crosslingual results form a different picture for Spanish and Basque. The Spanish model trained in the crosslingual transfer setting (en2es) scores particularly well on Richness, Coherence and Grammaticality, and outperforms the data-transfer version es-mt for all five criteria. However, for Basque, the model-transfer setting (en2eu) is particularly low in terms of Relatedness and Specificity. Still, the overall scores of en2eu are better than those of eu-mt. In any case, it should be noted that for Basque the en2eu model-transfer approach slightly outscores the data-transfer method (eu-mt), the only result that does not correlate with the automatic evaluation discussed in the previous section.

\paragraph{Correlation} Summarizing, the results for Spanish from both quantitative and qualitative evaluations correlate for every evaluation setting. This however, is not the case for Basque as the zero-shot crosslingual transfer (en2eu) is better according to the manual evaluation exercise but substantially worse in terms of the automatic metrics. In the next section we provide an error analysis to better understand this and other issues related with the automatic and the manual evaluation of CN generation.

\section{Discussion}\label{sec:error}

An error analysis was conducted in order to examine the quality of the generated CNs. In this section, we focused
our attention on the Basque generated outputs. As previously mentioned, although eu-post performed surprisingly well in the manual evaluation, it still produces cases for which the scores in Relatedness and Specificity are rather low. An example of an unrelated and generic CN generated by eu-post is illustrated in Example \ref{bad-eu-example}. We can observe how the generated CN is not replying to the HS: while the HS is asking for the repatriation of Muslims, the CN is questioning why Muslims cannot build mosques in their country.


\begin{example} \label{bad-eu-example}

\mbox{}
\begin{itemize}
\item[\textbf{HS}] Musulmanak gure herrialdetik atera beharko lirateke inork ez dituelako onartzen.

`Muslims should be taken away from our country because they aren't accepted by anyone.' 

\item[\textbf{CN}] Zergatik ez zaie baimendu behar musulmanei meskita hemen eraikitzea? Herrialde honek kristautasunaz gain beste erlijio batzuk praktikatzeko askatasuna ematean sinesten zuela uste nuen.

`Why should not Muslims be allowed to build Mosques here? I thought this country believed in giving people the freedom to practice religions other than Christianity?' 
\end{itemize}

\end{example}

A very similar phenomenon occurs when we look at the outputs generated by mT5 when fine-tuned in a zero-shot crosslingual setting (en2eu). Table \ref{tab:cross-results} shows that en2eu obtained the worst overall scores across all models. However, en2eu performs considerably better in the manual evaluation, obtaining better overall results than eu-mt. After a closer analysis of the manual evaluation, it can be observed that although en2eu obtained the highest scores among the three Basque models for Richness and Grammaticality, its results for Relatedness and Specificity remain extremely low. Thus, even though mT5 generates grammatical CNs in an en2eu zero-shot crosslingual setting, they clearly fail to comply with the more domain-oriented human evaluation criteria (Relatedness and Specificity), as illustrated by Example \ref{bad-en2eu-example}.

\begin{example} \label{bad-en2eu-example}
\mbox{}
\begin{itemize}
\item[\textbf{HS}] Since 2000 all suicide bombings were committed by a Muslim and no one handle this fact that about global terrorism.

\item[\textbf{CN}] Musulman britainiarrek beste edozein talde erlijiosok baino gehiago ematen diote karitateari per capita [esteka MEND-eko estatistiketara edo/eta meskita bateko ongintzako dohaintza handien adibidera], beraz, kosta egiten zait ikustea funts horiek biltzen diren meskitekin zer arazoa dagoen.

`British Muslims give more to charity, per capita, than all other religious groups [link to mend statistics and/or example of one large charity donation from a Mosque], so I am struggling to see the problem with Mosques where these funds are mostly raised.' 

\end{itemize}
\end{example}

This example illustrates how the generated response, while perfectly fine in terms of grammaticality and richness, is neither related nor specific to the HS it is supposed to be responding to.

Finally, a note about the coherence of the generated CNs. Despite of the automatic evaluation, human evaluation shows that Coherence results are consistently high across all six evaluated settings. Therefore, our manual evaluation shows that the main problems of our approach to generate CNs are related with the Relatedness and Specificity criteria, and not so much with Coherence and Grammaticality.

In this sense, while we have followed the evaluation methodology from previous works using the original CONAN dataset, further research on the automatic evaluation of CN generation is needed as current automatic overlap-based metrics fail to capture application-specific features or aspects of the generated CNs.

\section{Concluding Remarks}\label{sec:conclusion}

In this paper we have presented CONAN-EUS, a new parallel Basque and Spanish dataset for CN generation consisting of automatic translations and professional post-editions of the original English CONAN. The corpus consists of 6654 machine translated HS-CN pairs and 6654 gold-standard human-curated HS-CN pairs (per language) which makes it a unique resource to investigate CN generation from a multilingual and crosslingual perspective. 

Experimental results show that CN generation is better when mT5 is fine-tuned on post-edited training data, rather than on the output of MT. Multilingual experiments (training on all three languages) show that crossligual transfer works best for English and Spanish, which could be due to a number of reasons: (i) English and Spanish and structurally more similar than with respect to Basque and, (ii) the quota of shared strings in the mT5 vocabulary is larger for those languages than for Basque, a less-resourced language. Evaluation on zero-shot crosslingual settings produced similar results. While model-transfer outperforms data-transfer for Spanish, the opposite is true for Basque. Still, crosslingual transfer in generative approaches remains an open and complex research problem \cite{lin-etal-2022-shot}. 

Our work underpins several opportunities for future work on many domains, such as online hate mitigation targeting Basque and Spanish as well as crosslingual transfer with the aim of not only improving current techniques for CN generation, but also mitigating the lack of manually curated training data for many languages. For instance, while our experiments indicate that post-edited training data is required to attain reasonable performance, it is unclear how much additional such data would be a minimum for outperforming models trained on machine translated data. As manual annotation is costly and almost unattainable for gathering CN data, lessening the data requirement on expert effort is desirable. Furthermore, our work focuses on two languages as a first step towards examining data transfer and model transfer for CN generation. Future work is needed to investigate whether similar findings would hold in other languages or to pinpoint best practices for generating CN under low-resourced scenarios. 

\section{Acknowledgements}

This work has been partially supported by the Basque Government (Research group funding IT-1805-22). We are also thankful to the following MCIN/AEI/10.13039/501100011033 projects: (i) DeepKnowledge (PID2021-127777OB-C21) and by FEDER, EU; (ii) Disargue (TED2021-130810B-C21) and European Union NextGenerationEU/PRTR; Rodrigo Agerri currently holds the RYC-2017-23647 fellowship (MCIN/AEI/10.13039/501100011033 and by ESF Investing in your future). Yi-Ling Chung is supported by the Ecosystem Leadership Award under the EPSRC Grant EPX03870X1 \& The Alan Turing Institute.

\section{Ethical Statement}

Tasks on Counter Narrative generation and corresponding datasets have been proposed by the scientific community to try to mitigate the spread of online hate. However, even good and legitimate purposes may imply certain risks, especially for people involved in the development of datasets which may include extreme forms of Hate Speech. This includes researchers, annotators, data curators and evaluators involved in studies such as the one we present in this paper. Conscious of these facts, we took every precaution we could think of to avoid harmful effects.

\section{Bibliographical References}\label{sec:reference}

\bibliographystyle{lrec-coling2024-natbib}
\bibliography{lrec-coling2024-example}

\end{document}